\documentclass[letterpaper]{article}
\usepackage{aaai}
\usepackage{times}
\usepackage{helvet}
\usepackage{courier}
\usepackage{amssymb, amsmath ,color, stmaryrd, graphicx, algorithmic, algorithm, url}

\nocopyright

\begin{document}
%
\title{Experiments with Game Tree Search in Real-Time Strategy Games}
\author{Santiago Onta\~{n}\'{o}n \\
Computer Science Department \\
Drexel University \\
Philadelphia, PA, USA 19104 \\
santi@cs.drexel.edu }

\maketitle
\begin{abstract}
\begin{quote}
Game tree search algorithms such as minimax have been used with enormous success in turn-based adversarial games such as Chess or Checkers. However, such algorithms cannot be directly applied to real-time strategy (RTS) games because a number of reasons. For example, minimax assumes a turn-taking game mechanics, not present in RTS games. In this paper we present RTMM, a real-time variant of the standard minimax algorithm, and discuss its applicability in the context of RTS games.  We discuss its strengths and weaknesses, and evaluate it in two real-time games.
\end{quote}
\end{abstract}

\section{Introduction}\label{sec:intro}

Game tree search algorithms such as minimax have been used with enormous success in turn-based adversarial games such as Chess or Checkers. However, such algorithms cannot be directly applied to real-time strategy (RTS) games due to four main reasons \cite{buro2003rts,Aha05learningto}: 1) RTS games are real-time, 2) Game trees for RTS games have very large branching factors, 3) some RTS games are only partially observable, and 4) some RTS games are non-deterministic.

Some of these problems have been addressed in the context of game tree search. For example, sampling-based tree search algorithms like UCT \cite{Kocsis06banditbased} address the issue of having a large branching factor. In this paper we will focus on the first problem (the real-time nature of RTS games), and study it in the context of game tree search algorithms. Addressing the other three problems (large branching factors, partial observability, non-determinism) is out of the scope of this paper. This is a deliberate choice, since it is easier to study the real-time problem in isolation, and the ideas presented in this paper straightforwardly apply to other approaches that deal with large branching factors or with non-determinism.

Specifically, in this paper we study a real-time version of the minimax algorithm that we call {\em real-time minimax} (RTMM), and discuss its applicability to RTS games. There are two key problems that need to be addressed: a) how to represent a game tree when players do not take turns, but execute (potentially simultaneous) actions in real-time, and b) how to deal with the problem that the game keeps advancing while the AI is spending time searching the game tree. 

The rest of this paper is organized as follows. First we present some preliminary background on game tree search, followed by a description of the RTMM algorithm. After that, we discuss how RTMM can be deployed in the context of a RTS game. Finally, we present an empirical evaluation of the algorithm in two real-time games.

\section{Background}\label{sec:background}


A classic turn-taking game, such as Chess can be defined as a tuple $G = (S, A, P, T, V, W, s_{init})$, where:

\begin{itemize}
\item $S$ is the set of possible states in the game (e.g. in Chess, the set of all possible board configurations).
\item $A$ is the finite set of possible actions that can be executed in the game (e.g. ``move a pawn from B2 to B3''). 
\item $P$ is the set of players (in this paper, we will assume that there are only two players, i.e.  $P = \{min,max\}$).
\item $T : S \times A \to S$ is the deterministic transition function, that given a state and an action, returns the state resulting from applying the given action to the given state.
\item $V : S \times A \times P \to \{true,false\}$ is a function that given a state, an action and a player, determines if the given player can execute the given action in the given state.
\item $W : S \to P \cup \{draw, ongoing\}$ is a function that given a state determines the winner of the game, if the game is still ongoing, or if it is a draw.
\item $s_{init} \in S$ is the initial state.
\end{itemize}

In order to apply game tree search, an additional {\em evaluation function} is typically provided, The evaluation function predicts how attractive is a given state for a given player. 
We will assume an evaluation function of the form 
$E : S \to \mathbb{R}$, which returns positive numbers for states that are good for $max$ and negative numbers for states that are good for $min$. 

\begin{algorithm}[tba] \caption{minimax(s, d, p)}\label{alg:minimax} \begin{algorithmic}[1]
\medskip
\IF {$d \leq 0 \,\, \vee \,\, W(s) \neq \mathit{ongoing}$}
\RETURN $E(s)$
\ENDIF
\IF {$p = max$}
\STATE {$b = -\infty$}
\FORALL {$a \in A$ such that $V(s,a,max) = true$}
\STATE {$b = max(b,\text{minimax}(T(s,a), d-1, min))$}
\ENDFOR
\RETURN $b$
\ELSE 
\STATE {$b = \infty$}
\FORALL {$a \in A$ such that $V(s,a,min) = true$}
\STATE {$b = min(b,\text{minimax}(T(s,a), d-1, max))$}
\ENDFOR
\RETURN $b$
\ENDIF
\end{algorithmic}
\end{algorithm}

Using this notation, Algorithm \ref{alg:minimax} shows the standard minimax algorithm, which, coupled with $\alpha$-$\beta$ pruning \cite{knuth1975alphabeta} is the most common game tree search algorithm. Notice that the algorithm assumes turn-taking, one action per player per turn, and instantaneous actions. 

There has been some work on extending game tree search ideas with the goal of handling RTS games. For example, in domains where players can execute simultaneous actions, minimax is well known to under or over estimate the value of positions. Approaches like SMAB (Simultaneous Move $\alpha$-$\beta$) \cite{SaffidineFinnssonBuro2012AAAI} or {\em randomized $\alpha$-$\beta$} \cite{KovarskyB05heuristic} address this problem, and bring minimax search one step closer to be applicable in RTS games, where simultaneous actions are allowed. Although these approaches still focus on turn-based games, their ideas can be applied directly to the algorithm presented in this paper, as we will see later, in the experimental evaluation section.

Chung et al. \shortcite{ChungBS05} studied the applicability of game tree Monte Carlo simulations to RTS games, and proposed the MCPlan algorithm. MCPlan uses {\em high-level plans}, where a plan consists of a collection of destinations for each of the units controlled by the AI. MCPlan generates a collection of random high-level plans, and then simulates them multiple times pitting them against a sample of the possible plans the opponent can perform. At the end of each simulation, an evaluation function is used, and the plan that performed better overall is selected. 
Since plans take longer to execute than lower-level actions, this reduces the amount of search needed to play the game, making the approach computationally feasible in practice. The idea was continued by Sailer et al. \cite{SailerBL07} where they studied the application of game theory concepts to MCPlan. 

MCPlan uses the notion of a ``plan'' to abstract away from the RTS game underneath and make it amenable to game tree search techniques. In this paper, however, we study how to apply game tree search directly to real-time games without using abstraction. Although we believe abstraction will be necessary to deal with commercial RTS games (like Starcraft), in this paper we make the deliberate choice of not using it, since we want to consider the simplest possible scenario to study game tree search in real-time domains.


A more closely related work to RTMM is that of Balla and Fern \shortcite{balla2009uct}, where they study the application of the UCT algorithm \cite{Kocsis06banditbased} (a Monte Carlo game-tree search algorithm) to the particular problem of tactical battles in RTS games with great success. In their work, they use abstract actions that cause groups of units to merge or attack different enemy groups. Balla and Fern deal with the problem of searching in game trees with durative actions, but they do not deal with the problem of performing such search under real-time constraints (where the game keeps running in parallel while the agent is performing search). In their experiments, the game was slowed down so that the game tree could be expanded each time a new action had to be issued. Another application of UCT to real-time games is that of Samothrakis et al. \shortcite{Samothrakis2011mspacman}, in the game Ms. Pac-Man, where they first re-represent Ms. Pac-Man as a turn-based game, and then apply UCT.

Finally, we would like to point out that even if there are algorithms that use minimax ideas for real-time search, such as Min-max LRTA* \cite{Koenig01minimaxrta}, those are not designed for adversarial search, but are relatives of the A* algorithm.

In this paper we address two problems: defining game trees for real-time domains, and determining when to spend time to search the game tree while the game is running. The next two sections deal with these two problems.

\section{Game Trees for Real-Time Domains}\label{sec:framework}



The dynamics of RTS games cannot be defined with a transition function of the form $T:S \times A \to S$ as in the case of turn-based games since actions may be durative and also occur simultaneously. Therefore, we need to modify the transition function accordingly.

In RTS games, players control a collection of individual units that players issue actions to. Each of these units is associated with a player and can only execute one action at a time, but, since there might be multiple units in a game state, players can issue multiple actions at the same time (one per unit).  We will refer to those actions as {\em basic-actions}. 

To account for simultaneous actions, the function that determines which actions can be executed needs to be redefined as: $V : S \times 2^A \times P \to \{true,false\}$.



We will define a {\em player-action} $\alpha$ as a set of basic-actions that can be executed simultaneously: $\alpha = \{a_1, ..., a_n\}$. 
We will use lower case $a$ for basic-actions, and the greek letter $\alpha$ to denote player actions.
Given a game state $s$, we define $\mathit{PlayerActions}(s,p) = \{\alpha \subseteq A \mid V(s,\alpha,p) = true\}$ as the set of possible player-actions the player $p$ can issue in the game state $s$. When $\mathit{PlayerActions}(s,p) = \{\emptyset\}$, i.e., the only available player action is the empty action, we say that player $p$ cannot issue any action in the game state $s$. Notice that it is impossible that $\mathit{PlayerActions}(s,p) = \emptyset$, since, if there is any unit that can execute an action, then $\mathit{PlayerActions}(s,p) \neq \emptyset$, and if there is no unit that can execute any action, then, by definition of player-action, $\mathit{PlayerActions}(s,p) = \{\emptyset\}$.

\subsection{Time}

In classical turn-based games, time advances each time a player issues an action. However, in real-time games, time progresses independently from the actions that are issued by the players. We will consider that game states have a time stamp, $time(s)$, and 
define the dynamics of real-time games by the following functions:

\begin{itemize}
\item $issue : S \times 2^A \times P \to S$, that given a game state, a player action, and a player, returns the corresponding game state where those actions have been issued. Notice that if $s' = issue(s,\alpha,p)$, then $time(s) = time(s')$, i.e. issuing an action does not advance time.
\item $simulate : S \times \mathbb{R}^+ \to S$, that given a game state $s$ and a time $t$, runs the game from $s$ until any of the two players can issue a new action, the game ends, or until time $t$ has been reached. Thus, if $s' = simulate(s,t)$, we know that $time(s) \leq time(s') \leq t$. This captures the intuition that the game state only changes when both players have issued their actions.
\end{itemize}

Moreover, we assume that each {\em basic-action} $a$ takes a finite and deterministic amount of time. The amount of time left for an action $a$ in a given game state $s$, is given by the function: $\mathit{ETA}: A \times S \to \mathbb{R}^+$. Also, we assume that actions cannot be interrupted. Once an action has been issued, players must wait until it completes execution before another basic-action can be issued to the same unit. 

Notice that this model of basic-actions is less restrictive than it seems. RTS games typically contain complex actions such as ``harvest'', that can be interrupted, and for which it is hard to define a deterministic execution time.
In our model, we don't consider actions like ``harvest'' to be basic-actions, but as macro-actions; i.e., as a convenience so that the human player does not have to micro manage the individual unit. In fact, such macro-action is composed of a long sequence of individual basic-actions, such as: ``move one cell to the right''. 

Finally, notice that, in the context of a RTS game, the function $simulate$ as defined here, would not let time advance unless each individual unit in the game is executing a basic-action.  When a unit completes an action, the player must issue another action to that unit immediately. Notice that this is not restrictive in practice, since we can have a special action {\em no-action} defined as taking a small amount of time for those situations when players do not want to issue any particular action to a given unit. This is the approach taken in the experiments reported in this paper.

%
%

\subsection{Real-Time Minimax}\label{sec:rtminimax}

With the previous definitions, let us present the basic form of the real-time minimax (RTMM) algorithm. In this section, we will ignore the fact that the game keeps changing while the AI spends time expanding the tree (we will deal with this issue in the next section). Moreover, as we discussed earlier, we will focus on the 2 player scenario, although generalizing to more than 2 players can be done by using the same idea as in the $\mathit{max}^n$ algorithm \cite{LuckhartI86maxn}.

\begin{algorithm}[tba] \caption{RTMM(s, $t_{max}$)}\label{alg:rtminimax} \begin{algorithmic}[1]
\medskip
\IF {$time(s) \geq t_{max} \,\, \vee \,\, W(S) \neq \mathit{ongoing}$}
\RETURN $E(S)$
\ENDIF
\IF {$\mathit{PlayerActions}(s,max) \neq \{\emptyset\}$} 
\STATE {$b = -\infty$}
\FORALL {$\alpha \in \mathit{PlayerActions}(s,max)$}
\STATE {$b = max(b, \text{RTMM}(issue(s,\alpha,max),t_{max})$)}
\ENDFOR
\RETURN $b$
\ELSIF {$\mathit{PlayerActions}(s,min) \neq \{\emptyset\}$} 
\STATE {$b = \infty$}
\FORALL {$\alpha \in \mathit{PlayerActions}(s,min)$}
\STATE {$b = min(b, \text{RTMM}(issue(s,\alpha,min),t_{max})$)}
\ENDFOR
\RETURN $b$
\ELSE
\RETURN {RTMM$(simulate(s, t_{max}), t_{max})$}
\ENDIF
\end{algorithmic}
\end{algorithm}

Algorithm \ref{alg:rtminimax} presents the RTMM algorithm. RTMM only takes two input parameters: $s$ is the current state of the game, and $t_{max}$ is the time up to which we want to open the game tree. Comparing it to the original miminax algorithm, 
there are three main differences:

\begin{itemize}
\item RTMM cuts off search by  time $t_{max}$ rather than depth.
\item RTMM doesn't need the parameter specifying which player is next to move since in real-time games, any player can move at any time (they can even move simultaneously). Instead of looking who is the next player to move, RTMM checks if any player can issue any action at any given time (by checking if $\mathit{PlayerActions}(s,max)$ or $\mathit{PlayerActions}(s,min)$ are different from $\{\emptyset\}$). Notice that this means that, unlike in the standard version of minimax, {\em max} and {\em min} layers do not necessarily alternate.
\item In case no player can issue any action, RTMM has a third case (line 17), where the game is simulated until a player can issue some action (using the {\em simulate} function).
\end{itemize}



\begin{figure*}[ta]
    \centering
    \includegraphics[width=13cm]{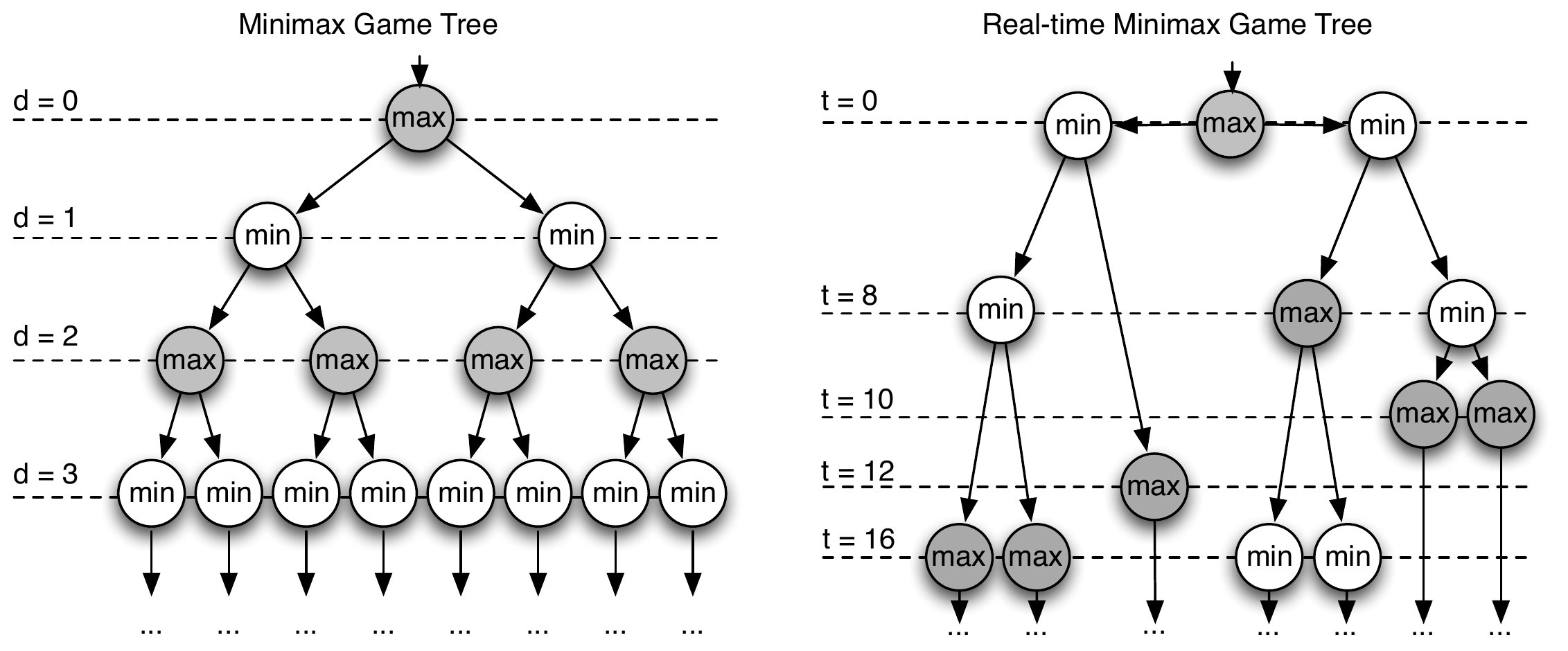}
    \caption{A game tree generated with standard minimax (left), and a game tree generated with RTMM (right).}
    \label{fig:tree-comparison}
\end{figure*}

In order to compare the game trees that RTMM generates with the game trees generated by standard minimax, Figure \ref{fig:tree-comparison} shows game trees generated by both algorithms side by side. The minimax game tree has been sorted by depth ($d$) of the nodes (deeper nodes appear lower in the figure), and the RTMM game tree has been sorted by time ($t$) of the nodes (nodes corresponding to game states with larger time stamps appear lower in the figure). While nodes in standard minimax game trees are organized in a series of alternating layers (all the nodes at depth 1 are min nodes, all the nodes at depth 2 are max nodes, etc.), that is not true for the RTMM game trees. In the game trees expanded by RTMM min and max nodes can appear in any order, since that is the nature of real-time games. There can even by time stamps, for which there are simultaneous max and min nodes (such as in time $t = 0$ in the figure).

Moreover, notice that sometimes both players can execute moves simultaneously, like at time $t = 0$ in the right hand side of Figure \ref{fig:tree-comparison}. In those situations, RTMM simply, first considers all the moves for the {\em max} player, and then all the moves for the {\em min} player (i.e. it does a {\em max-min} strategy. This might lead to an overestimation of the value of certain nodes. However, there are existing techniques in the literature that can be applied on top of RTMM to address this situation, if simultaneous moves is a significant part of the game at hand (for example \cite{KovarskyB05heuristic}).

\section{Application to Real-Time Strategy Games}\label{sec:rts}


This section focuses on how to use game trees in the context of RTS games, where the game keeps advancing while the AI might be expanding the game tree.


Computer games, and RTS games in particular, run by executing a series of {\em game cycles}. At each cycle, the game state is updated, and each player has a chance to issue actions. Typically, games execute about 50 cycles per second, which means that the AI has only a fraction of a second to issue actions after receiving the updated game state.


Therefore, we will assume that the AI is called once per game cycle with the updated game state $s$, and given a predefined amount of time $T$ to execute before it has to issue actions. Also, we will assume that we can run RTMM for a certain amount of time, interrupt it, and resume it later if need be.
Using this idea, an AI system using the RTMM algorithm to play a RTS game, should act like this (we assume the AI controls player {\em max}):

%


\begin{itemize}
\item If we are not the next player that will have a chance to issue an action (i.e. $\mathit{PlayerActions}(\mathit{simulate}(s,\infty)) = \{\emptyset\}$), then do nothing.
\item If we are the next player that will have a chance to issue an action, but we cannot issue any action in the current cycle, then: if there was a search already started from the previous game cycle, continue it, otherwise, start a new search from state $s' = \mathit{simulate}(s,\infty)$ (that is the state in which we will be able to issue the next action).
\item If we are the next player that will have a chance to issue an action, and it can be issued in the current cycle, then continue or start a new search, and return the best action.
\end{itemize}

Two things need to be highlighted: 1) the AI might be able to use several consecutive frames in a row to expand deeper search trees, and 2) new game tree searches are not started from the current game state $s$, but from $s' = \mathit{simulate}(s,\infty)$ (i.e. to the point in time when we need to issue an action).

In our experiments we observed that there are many consecutive cycles where the AI is just waiting for actions to complete execution, and thus there is often a significant amount of time available for game tree search.

\subsection{Iterative Deepening}\label{sec:deepening}

Typically, when minimax is given a certain amount of time to run, an iterative deepening technique is used, where the tree is first expanded to depth 1, then to depth 2, etc. until time runs out. Since the concept of tree depth in RTMM doesn't make sense (RTMM is given a maximum time $t$, not a depth $d$), the idea of iterative deepening has to be modified.

Let us define $F(s,t) = \{s_1, ..., s_n\}$ as the set of game states in the leaves of the tree that would be explored by the execution of RTMM using a cut-off time $t$ from game state $s$. Some of those game states will have been reached because the simulation reached the cut-off time $t$, and some others because the simulation reached the end of the game. Let us define $F^*(s,t) \subseteq F(s,t) $ as those game states that were reached because the simulation reached the cut-off time, and:
$$t^- = min_{s \in F^*(s,t)} time(simulate(s,\infty))$$
$$t^+ = max_{s \in F^*(s,t)} time(simulate(s,\infty))$$
Clearly $t^+ \geq t^- \geq t$. This means that, we know that if we repeat the execution of RTMM with a larger cut-off time $t'$ such that $t \leq t' \leq t^-$, the tree that will be explored is identical to that explored with $t$. Thus, if we want to perform iterative deepening, we need to run the next iteration of the algorithm with a cut off time $t'$ such that $t' > t^-$. Moreover, we know that if we re-run the algorithm with $t' \geq t^+$, every single branch that did not reach the end of the game will be expanded with at least one max or one min layer. 

As a conclusion, for performing iterative deepening, given that we have used a time $t$ for one iteration, the next iteration should be run with $t' \in (t^-,t^+]$. Through empirical validations, we observed that $t' = t^- + \epsilon$ (where $\epsilon$ corresponds to the length of the shortest of all basic-actions defined in the RTS game) woks best in practice.

%
%
%


\subsection{Extensions of the Basic Algorithm}

The RTMM algorithm, as presented in in this paper is intended to be a theoretical algorithm, with the sole purpose of illustrating that it is possible to open  game trees in real-time domains. In order to make the RTMM scale up to domains of practical relevance, it can be extended in many ways.

Enhancements such as $\alpha$-$\beta$ pruning \cite{knuth1975alphabeta} (used in our experiments), transpositions tables \cite{atkin1988transposition}, the history heuristic \cite{Schaeffer89history}, or opening books can be trivially applied.
Additionally, the key idea behind RTMM can be applied to other algorithms such as UCT \cite{Kocsis06banditbased}, in order to use UCT in real-time domains. Finally, for the sake of simplicity of presentation, in this paper we have focused in deterministic domains. However, non-deterministic domains can be handled by RTMM by turning the simulation layers into {\em averaging} layers (like in the expectiminimax algorithm \cite{russell2009ai}) used for games like Backgammon.

Finally, RTMM is based on minimax, and therefore, as mentioned earlier, it tends to over or underestimate the value of positions when the two players can perform moves simultaneously. Many of the solutions that have been proposed to deal with this problem, such as randomized $\alpha$-$\beta$ \cite{KovarskyB05heuristic} are directly applicable to RTMM. In fact, in our empirical evaluation below, we experimented with a modified version of RTMM that uses randomized $\alpha$-$\beta$.

\section{Experiments}

In order to evaluate RTMM we used two open-source real-time games: BattleCity \cite{ontanon2009d2learning} and $\mu$RTS. In both domains, we implemented the $\alpha$-$\beta$ pruning variant of RTMM with iterative deepening, and evaluated it against other AIs. 

Additionally, we also tested the performance of RTMM with the addition of randomized $\alpha$-$\beta$ \cite{KovarskyB05heuristic} (we refer to this as RRTMM below). In order to do this, we just incorporated the main idea of randomized $\alpha$-$\beta$ on top of the RTMM algorithm, i.e. for all internal nodes of the tree, other than the head, whenever there is a simultaneous move situation, we randomly pick whether we would search that node using a max-min or a min-max strategy.

\subsection{BattleCity}

BattleCity is a recreation of the original BattleCity game by Namco, where each player controls a tank through a PacMan-like labyrinth. Each player needs to defend its base (a static square) while trying to destroy the other player's tank or base. One interesting strategical feature is that some walls can be destroyed. Players can move the tank 
in any of the 4 directions (up, down, left, right) and shoot. Shooting and movement are independent, and thus, players can issue up to two actions simultaneously. The player can fire once each 8 cycles, and move the tank once each 16 cycles. The game is deterministic, and fully observable.

For testing in BattleCity, we used two of the AIs that are built-in into the game: {\em Random} (which moves at random and fires constantly unless its base is in the line of fire), and {\em Follower} (which uses $A^*$ to find the closest path to either the closest enemy tank or base and shoots constantly unless its base is in the line of fire). As a reference, the {\em Random} is easily beaten by a human, but humans struggle to defeat {\em Follower} unless the map allows them to set up an ambush.

All the reported results are the result of 10 executions on 6 different maps varying in sizes from 18x18 to 26x18 tiles. All games were cut-off after 5000 cycles if there was no winner (considering it a tie). The evaluation function used for RTMM was very simple: -$\infty$ for loss, $\infty$ for win, and manhattan distance of the enemy to our base minus manhattan distance of our tank to the enemy's base for any other case (i.e. the heuristic pushes the tank towards the enemy base).

\begin{table}[tba]
\centering
\small
\begin{tabular}{|l|c|c|c|c|} \hline
				& {\em Random} 	& {\em Follower} 	& {\em RTMM}	& {\em RRTMM} \\ \hline
{\em Random} 		& 30-0-30 	& 40-3-17 	& 54-0-6 		& 50-1-9 \\ \hline
{\em Follower} 		& 17-3-40 	& 5-50-5 		& 51-0-9 		& 50-0-10 \\ \hline
{\em RTMM} 		& 6-0-54 		& 9-0-51 		& 10-40-10 	& 9-38-13 \\ \hline 
{\em RRTMM} 		& 9-1-50 		& 10-0-50 	& 13-23-9 	& 11-38-11 \\ \hline \hline
{\em total}			& 62-4-174 	& 64-53-123	& 128-78-34 	& 120-77-43 \\ \hline
\end{tabular}
\caption{Win/tie/lose statistics for the four bots tested in the BattleCity domain, for each bot in each column we show wins-ties-loses against the bot in the corresponding row.}
\label{tbl:battlecity}
\end{table}

Table \ref{tbl:battlecity} shows the number of wins, ties and loses that each different bot obtained in the different match-ups. We can see that, both bots using RTMM are vastly superior to both the Random bot (against which RTMM won 54 out of 60 times) and the Follower bot (against which RTMM won 51 out of 60 times). In fact, we observed that when the tanks are close together, RTMM plays almost optimally, since it can open the search tree deep enough. However, the further away the tanks are, the more RTMM has to rely on the evaluation function and its level of play degrades. This shows that RTMM with a very simple heuristic is enough to play real-time games like BattleCity at a high level of game play. The randomized $\alpha$-$\beta$ modification of RTMM (RRTMM) didn't improve the performance (it actually slightly decreased). The reason is that, even if randomized $\alpha$-$\beta$ has been reported to obtain significant gains in simultaneous move games \cite{KovarskyB05heuristic}, the effect of simultaneous moves in BattleCity is minimal, but results in the bot being more aggressive, which results in more losses. Moreover, as we will see in the next section, randomized $\alpha$-$\beta$ search improved the performance of RTMM on the other domain we tested.

The first column of Table \ref{tbl:stats} shows the minimum and maximum time that RTMM could spend in expanding each of the game trees in BattleCity, showing that it never had more than 80ms (which accounts for being able to continue expanding the tree for 8 frames in a row). The branching factor of the game tree in BattleCity is between 2 to 8, and in our experiments, RTMM opened rather large trees, with up to 248,400 leaves \footnote{Notice that this number might look small in the context of standard game tree search, but we must have in mind that these trees are opened in real-time, in only a fraction of a second.}, being able to look ahead up to 1.44s in game time. We would like to emphasize that no simplifications of the game were employed to apply RTMM, which plays the game with all the complexity that a human would.

\begin{table}[tba]
\centering
\small
\begin{tabular}{|l|c|c|c|} \hline
& {\em BattleCity} & {\em $\mu$RTS-Melee} & {\em $\mu$RTS-Full} \\ \hline
Search Time & 10 - 80ms & 100ms - 1.2s & 100ms - 1s \\ \hline
Branching & 2 - 10 & 4 - 600 & 4 - 76560 \\ \hline
Max Leaves & 248400 & 150525 & 207015 \\ \hline
Max Depth & 1.44s & 7.20s & 4.90s \\ \hline
\end{tabular}
\caption{Game Tree Search statistics for the domains used in our evaluation. Max depth is measured in seconds, since the depth of a RTMM tree is measured in time.}
\label{tbl:stats}
\end{table}

\subsection{$\mu$RTS}

\begin{table}[tba]
\centering
\small
\begin{tabular}{|l|c|c|c|c|} \hline
				& {\em Stochastic} 	& {\em Rush} 	& {\em RTMM} & {\em RRTMM} \\ \hline
{\em Stochastic} 	& 10-0-10 		& 20-0-0 		& 20-0-0 		& 20-0-0\\ \hline
{\em Rush} 		&   0-0-20 		& 0-20-0 		& 16-4-0 		& 19-1-0\\ \hline
{\em RTMM} 		&   0-0-20 		& 0-4-16 		& 0-20-0 		& 0-20-0\\ \hline \hline
{\em RRTMM} 		&   0-0-20 		& 0-1-19 		& 0-20-0 		& 0-20-0\\ \hline \hline

{\em total}			& 10-0-70 		& 20-25-35	& 36-44-0 	& 39-41-0\\ \hline
\end{tabular}
\caption{Win/tie/lose statistics for the four bots tested in the $\mu$RTS-Melee domain, for each bot in each column we show wins-ties-loses against the bot in the corresponding row.}
\label{tbl:rts-melee}
\end{table}

\begin{table}[tba]
\centering
\small
\begin{tabular}{|l|c|c|c|c|} \hline
				& {\em Stochastic} 	& {\em Rush} 	& {\em RTMM} & {\em RRTMM} \\ \hline
{\em Stochastic} 	& 10-0-10 		& 15-0-5 		& 18-0-2 		& 15-0-5\\ \hline
{\em Rush} 		&   5-0-15 		& 10-0-10 	& 0-0-20 		& 0-0-20\\ \hline
{\em RTMM} 		&   2-0-18 		& 20-0-0 		& 0-20-0 		& 0-20-0\\ \hline \hline
{\em RRTMM} 		&   5-0-15 		& 20-0-0 		& 0-20-0 		& 0-20-0\\ \hline \hline

{\em total}			& 22-0-58 		& 65-0-15		& 18-40-22 	& 15-40-25\\ \hline
\end{tabular}
\caption{Win/tie/lose statistics for the four bots tested in the $\mu$RTS-Full domain, for each bot in each column we show wins-ties-loses against the bot in the corresponding row.}
\label{tbl:rts-full}
\end{table}

$\mu$RTS is a simple implementation of a RTS game where the map is a $n\times m$ grid, and there are only 6 types of units: worker, base, barracks, light combat unit, heavy combat unit, and mine. Workers can get resources form the mine and build buildings, barracks can train combat units. Different units have different amount of hit points, movement speeds, attack speed and power. The game is completely deterministic, and there is full observability. The actions that can be performed are (for each applicable unit): move (in 4 possible directions), build (in a neighboring cell), get resources (from a neighboring mine), drop resources (to a neighboring base), train (new unit will appear in a neighboring cell), and attack (to an enemy unit in range). This gives each unit a maximum of  24 actions at any given time.

For testing in $\mu$RTS we used two AIs: {\em Stochastic} (which selects moves stochastically, with a strong bias towards attacking, producing units, or moving towards the enemy), and {\em Rush}, which performs a light combat units rush strategy. The evaluation function used for RTMM is very simple and basically adds the resource cost of all the friendly units and subtracts the resource cost of all the enemy units.

We evaluated RTMM in two different scenarios: a 4vs4 units melee situation in a 8x8 map ($\mu$RTS-Melee), and a full 8x8 game, where players just start with a worker and a base ($\mu$RTS-Full). Games that ran for longer than 5 minutes where considered a draw. Table \ref{tbl:rts-melee} shows that in the $\mu$RTS-Melee scenario, RTMM-based bots (and specially RRTMM, the randomized $\alpha$-$\beta$-based bot) excel, defeating the other AIs consistently. However, the situation changes in the full game scenario, as shown in Table \ref{tbl:rts-full}. In this scenario, RTMM bots manage to perform better than the Stochastic bot, but fail to defeat the Rush but. RTMM plays perfect short term micro-scale game, but plays a very bad high-level (long term) strategy, and thus, is always overpowered by the Rush bot, which implements a good high-level strategy.

The reason for which RTMM can defeat the other AIs easily in the melee scenario but not in the full game scenario can be seen in Table \ref{tbl:stats}: the branching factor in the full game scenario grows to such large numbers (tens of thousands!) that it is not possible to perform too much lookahead. Even if Table \ref{tbl:stats} shows that RTMM was able to reach a maximum lookahead of 4.9 seconds, that is only at the beginning of the game, when there are few units in the board. In the melee scenario, the effect of simultaneous moves is important, and thus the randomized $\alpha$-$\beta$ modification improves the results. However, in the full-game experiments, simultaneous moves have a very small effect compared to high-level strategy.

\section{Conclusions}

Game tree search techniques such as minimax are typically assumed inapplicable to RTS games. The main contribution of this paper is questioning such assumption and studying RTMM, a real-time variant of the classic minimax game tree search algorithm, in the context of RTS games. Specifically, in this paper we have focused on addressing the fact that RTS games are real-time. 
The ideas behind RTMM can be straightforwardly applied to techniques that address some of the other problems of RTS games, such as to Monte Carlo search techniques for large branching factors.


We have evaluated RTMM in two real-time games (BattleCity and $\mu$RTS) with positive results. We have seen that if the branching factor is not too high, RTMM can achieve a very high level of gameplay (as shown  in BattleCity or in the melee scenario of $\mu$RTS) without using any abstraction nor simplification in the game. In domains such as the full game scenario in $\mu$RTS, the problem of large branching factors plays a significant role, and thus, sampling-based game tree search techniques (such as UCT) would be needed. We would like to emphasize, that RTMM is not to be seen as an alternative to methods like Monte Carlo search, but as a framework to study game tree search in real-time games. As part of our future work, we will  apply the ideas presented in this paper to Monte Carlo search in order to address both the real-time nature of RTS games as well as their large branching factors. Additionally we would like to study the extension of RMM to non-deterministic domains.



\bibliography{References}
\bibliographystyle{aaai}

\end{document}